\documentclass[12pt]{article}

\usepackage{amssymb}
\usepackage{amsmath}
\usepackage{booktabs, tabularx}
\usepackage{graphicx}
\usepackage{bm}
\usepackage{amsfonts}%
\usepackage{mathtools}
\usepackage{hyperref}
\usepackage{multimedia}
\usepackage[affil-it]{authblk}
\usepackage{listings}
\usepackage{titlesec}
\usepackage[font=footnotesize]{caption}

\title{\Large Discovering Antagonists in Networks of Systems: \\ Robot Deployment} 
\author[1]{\mdseries Ingeborg Wenger} 
\author[1\footnote{Corresponding author. Email address: peter.eberhard@itm.uni-stuttgart.de}]{Peter Eberhard} 
\author[2]{Henrik Ebel} 

\affil[1]{Institute of Engineering and Computational Mechanics, University of  \protect\\ Stuttgart, Pfaffenwaldring 9, 70569 Stuttgart, Germany}
\affil[2]{Department of Mechanical Engineering, LUT University, Yliopistonkatu 34, \protect\\ 53850 Lappeenranta, Finland}
\date{}

\titleformat{\section}
  {\fontsize{13}{15} \bfseries}
  {\thesection}{10pt}{}
\titleformat{\subsection}
  {\fontsize{12}{0} \itshape}
  {\thesubsection}{10pt}{}
\renewenvironment{abstract}
  {\small
   \begin{center}
   \bfseries \abstractname\vspace{-.5em}\vspace{0pt}
   \end{center}
   \list{}{
	 \setlength{\leftmargin}{.5cm}%
	 \setlength{\rightmargin}{\leftmargin}%
   }%
   \item\relax}
  {\endlist}

\begin{document}
\maketitle
\vspace*{-40pt}
\begin{abstract}
A contextual anomaly detection method is proposed and applied to the physical motions of a robot swarm executing a coverage task.
Using simulations of a swarm's normal behavior, a normalizing flow is trained to predict the likelihood of a robot motion within the current context of its environment.
During application, the predicted likelihood of the observed motions is used by a detection criterion that categorizes a robot agent as normal or antagonistic.
The proposed method is evaluated on five different strategies of antagonistic behavior.
Importantly, only readily available simulated data of normal robot behavior is used for training such that the nature of the anomalies need not be known beforehand.
The best detection criterion correctly categorizes at least 80\% of each antagonistic type while maintaining a false positive rate of less than 5\% for normal robot agents.
Additionally, the method is validated in hardware experiments, yielding results similar to the simulated scenarios.
Compared to the state-of-the-art approach, both the predictive performance of the normalizing flow and the robustness of the detection criterion are increased.

\end{abstract}

\section{Introduction}
\label{sec:introduction}
Snail-inspired robot swarms~\cite{ZhaoEtAl24}, a force-based approach to object transport~\cite{RosenfelderEbelEberhard24}, empathetic decision processes controlling swarm behavior~\cite{SiwekEtAl23}, and the collaboration of quadrotors and wheeled robots~\cite{ChenEtAl24} are merely examples for the fascinating ideas and applications that arise in the field of swarm robotics.
However, real-world applications of distributed robot swarms are lagging behind \cite{SchranzEtAl20}.
According to \cite{SchranzEtAl20}, a major concern for industrial applications are requirements for reliability, safety, and security, which are particularly relevant for tasks that involve close contact with humans and for applications that concern critical infrastructures.
The safety and security risks that have been identified in previous literature include the presence of intruding agents~\cite{SchranzEtAl20, SargeantTomlinson18, HigginsTomlinsonMartin09}, the injection of misleading information in the swarm~\cite{SchranzEtAl20, SargeantTomlinson18}, the assurance of swarm mobility~\cite{HigginsTomlinsonMartin09}, and the adherence to energy constraints~\cite{HigginsTomlinsonMartin09}.
Still, most literature assumes that all involved agents are functional and cooperative~\cite{SargeantTomlinson20} with some research conducted in the area of fault detection~\cite{QinHeZhou14}, but only limited work is available for the detection of antagonistic behavior.

Within the previously published work on misbehaving agents, several studies target the issue of reaching consensus within a swarm~\cite{SaulnierEtAl17, CavorsiEtAl22, DoostmohammadianElAl21, BabaeiGhazvini2023, Zhang2021}, e.g., regarding the synchronization of the robots' velocities~\cite{SaulnierEtAl17}, or the occurrence of events of interest~\cite{CavorsiEtAl22}.
Antagonistic behavior affecting a consensus problem is characterized by the transmission of incorrect, inconsistent, or misguiding information due to data injection~\cite{DoostmohammadianElAl21, BabaeiGhazvini2023}, the presence of a Sybil attack~\cite{CavorsiEtAl22} or Byzantine attacks~\cite{SaulnierEtAl17, Zhang2021}.
The effect of Byzantine agents, which transmit manipulated messages to the swarm with the goal of disrupting the assignment or execution of tasks, is further investigated in~\cite{Deng2021}. 
Other research detects attacks on robotic systems by means of analyzing the system's network traffic and power consumption~\cite{BasanBasanNekrasov19}, and applies a vision-based approach to detect differences in the sensor data of swarm members~\cite{SalimpourEtAl23}. 

Contrary to the aforementioned literature, this paper focuses on a deployment setting in which the antagonistic behavior manifests in the physical motions of an agent but generally benefits from the correct and reliable communication of these motions in order to manipulate the behavior of the swarm.
The deployment scenario is motivated by a variety of use cases that involve a robot swarm covering an area of variable size.
Such use cases include search and rescue missions, the monitoring of oil leaks, agricultural tasks, the provision of signal coverage, and general surveillance tasks~\cite{SchranzEtAl20, HigginsTomlinsonMartin09}.
The present work investigates the scenario of a robot swarm surveilling an area.
The swarm has the objective of achieving optimal coverage by distributing its agents in the area and assigning to each agent a sub-region to monitor.
A strategically positioned antagonist might be able to take control over a region of interest and to prevent other agents from detecting and alerting against illicit activities or entities within this region.
Thus, the surveillance scenario serves as a particularly straightforward example for the relevance of antagonist detection.

Previous literature on the detection of anomalies in the physical motions of swarm agents investigates a supervised learning approach~\cite{WangShangSun11}.
A neural network is trained to use the swarm's position and formation errors in order to classify the robot motions as normal or anomalous.
The training data for the class of anomalous motion behavior is created there by disturbing the robot's control input during a formation task.
During evaluation, the method successfully identifies robots that cease to move, apply an incorrect velocity, or move into the wrong direction.
However, the approach's reliance on formation errors and the restriction to a fixed number of swarm robots reduces its applicability to the deployment scenario investigated in the present work.

Additionally, a common challenge in anomaly detection is that a usable amount of labeled, empirical data on anomalous behavior is rarely available and typically not representative of all relevant types of anomalies~\cite{ChandolaBanerjeeKumar09}.
Thus, while the occurrence of physically anomalous behavior presupposes a deviation from an agent's expected behavior, the extent of this deviation is generally unknown and might differ greatly depending on the underlying issue or the nature of the anomalous behavior.
Similar objections arise for the intrusion detection system proposed in~\cite{SargeantTomlinson20}, which depends on the availability of signatures of anomalous behavior.
Moreover, the present work assumes antagonistic, instead of merely anomalous, behavior.
A restriction of the detection approach to a predefined set of behavior could therefore be easily circumvented or exploited by an antagonistic agent~\cite{ChandolaBanerjeeKumar09}.

A related issue concerns the work in~\cite{FagioliniEtAl09}, which presupposes a set of known, shared rules for normal behavior.
While the simulation or observation of data on normal swarm behavior should be straightforward, when considering more complex, learned, emergent, or non-deterministic swarm behavior, the definition of fixed rules on normal motion behavior might be challenging.
For instance, during the task of deploying a swarm in an area, the categorization of normal behavior is complicated by the dependence of a robot's motion on the coverage area and the positions of other swarm agents.

Based on these considerations, in~\cite{WengerEbelEberhard24} a data-based approach is proposed in order to detect anomalies in the physical motion of robot agents during a deployment task.
Taking the positioning of the robot swarm within the deployment area into account, a neural network is trained to predict the likelihood of a robot's motion behavior.
This prediction can then be processed by different detection criteria that categorize an agent as normal or antagonistic.
Notably, this approach only requires the availability of data on the normal motion behavior of robots during a deployment task without restricting the deployment scenario to a specific coverage area or a fixed number of agents.
Additionally, unlike in~\cite{CandidoHutchinson09}, the method does not rely on a centralized control unit to detect anomalies.

The present work is an extension and enhancement of our previous findings~\cite{WengerEbelEberhard24}, which successfully distinguished normally behaving robots from two types of faulty agents and one type of antagonistic agent with a sensitivity and specificity of more than 90\%.
However, several issues of the network and the detection criterion proposed in~\cite{WengerEbelEberhard24} were identified.
As demonstrated in the following, these issues particularly affect the method's application to the detection of more advanced antagonistic strategies that are introduced in this paper.
Consequently, the present work investigates some shortcomings of the method described in~\cite{WengerEbelEberhard24}.

Novel contributions include the representation of the robot swarm's position and the motion behavior, the neural network architecture, and the detection criterion.
The proposed extensions allow now to significantly enhance the assessment of the robot's physical behavior compared to~\cite{WengerEbelEberhard24}.
Thus, this paper exceeds the current state of the art for the anomaly detection in the motions of a robot swarm during a deployment task.
The applicability of the proposed approach is demonstrated on several strategies of antagonistic behavior, both in simulated scenarios and in hardware experiments.
The code and data are available on \href{https://github.com/Institute-Eng-and-Comp-Mechanics-UStgt/AntagonisticAgents}{GitHub}. 

The paper is organized as follows. 
Section~\ref{sec:coverage} introduces the deployment scenario as well as different strategies of antagonistic behavior.  
Section~\ref{sec:approach} describes the anomaly detection approach and provides details on the proposed enhancements of the neural network and the detection criterion.
The setup of simulations and hardware experiments can be found in Section~\ref{sec:experiments}, while Section~\ref{sec:result} provides the evaluation of the detection approach.

\section{Antagonistic strategies during a surveillance task}
\label{sec:coverage}

\begin{figure}[htbp]
	\includegraphics[width=\textwidth]{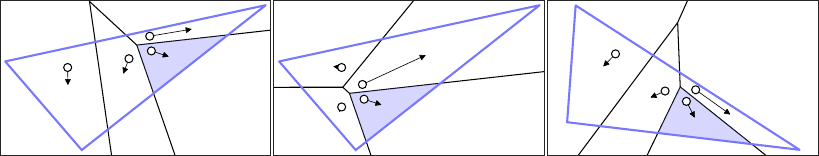}
	\caption[justification=left]{
		The Voronoi tessellation is used to determine the motion vectors of robots that exhibit normal behavior. The blue Voronoi cells show that the behavior of a robot might be invariant to the spatial translation of (local) objects, like neighboring robots, but not invariant to rotation within the inertial frame of reference.
		}
		\label{fig:voronoi}
\end{figure}

To carry out the surveillance task considered in this work, the robot swarm needs to distribute itself optimally within the deployment area.
There-by, in order to optimize the area's coverage, each point within the area should be as close as possible to one of the swarm robots' positions~\cite{ZhouEtAl24}.
The optimal robot positions can be found through an iterative process of robots selecting a new target position that depends on the current positions of the other agents, using Lloyd's~Algorithm~\cite{Lloyd82}.
At the beginning of each iteration, the robots communicate their current position to the swarm.
The position information received from other swarm members is subsequently used to construct the Voronoi tessellation of the deployment area, as visualized by the black lines in Figure~\ref{fig:voronoi}.
The Voronoi cell $W$ of a robot is defined as the set of all points in the area that are closer to the robot than to any other robot in the swarm~\cite{Voronoi08} and represents the region that a robot is expected to monitor.
To improve the monitoring of its assigned region, the robot selects a new target position within $W$.
This target position $\bm{x}^* \in \mathbb{R}^2$ is found by minimizing the distance from a candidate position $\hat{\bm{x}} \in \mathbb{R}^2$ to all points $\bm{q} \in \mathbb{R}^2$ within the robot's region $W$, i.e.,
\begin{equation}
	\bm{x}^* = \min_{\hat{\bm{x}}} \left[ \int_{W} ||\bm{q} - \hat{\bm{x}}||^2 \, \mathrm{d} W \right].
	\label{eqn:eqn_vor}
\end{equation}
Then, each robot moves towards its target before broadcasting its new position to the other robots within the swarm~\cite{Lloyd82}.
The position change $\bm{a} \in \mathbb{R}^2$ between two consecutive communicated robot positions is called an action.
As indicated by the colored Voronoi cells in Figure~\ref{fig:voronoi}, this action can be invariant to the translation of robot entities or the entire area within the inertial frame of reference, but not invariant to rotation.

\begin{figure}[htbp]
	\begin{center}
		\vspace*{-15pt}
		\includegraphics[width=\textwidth]{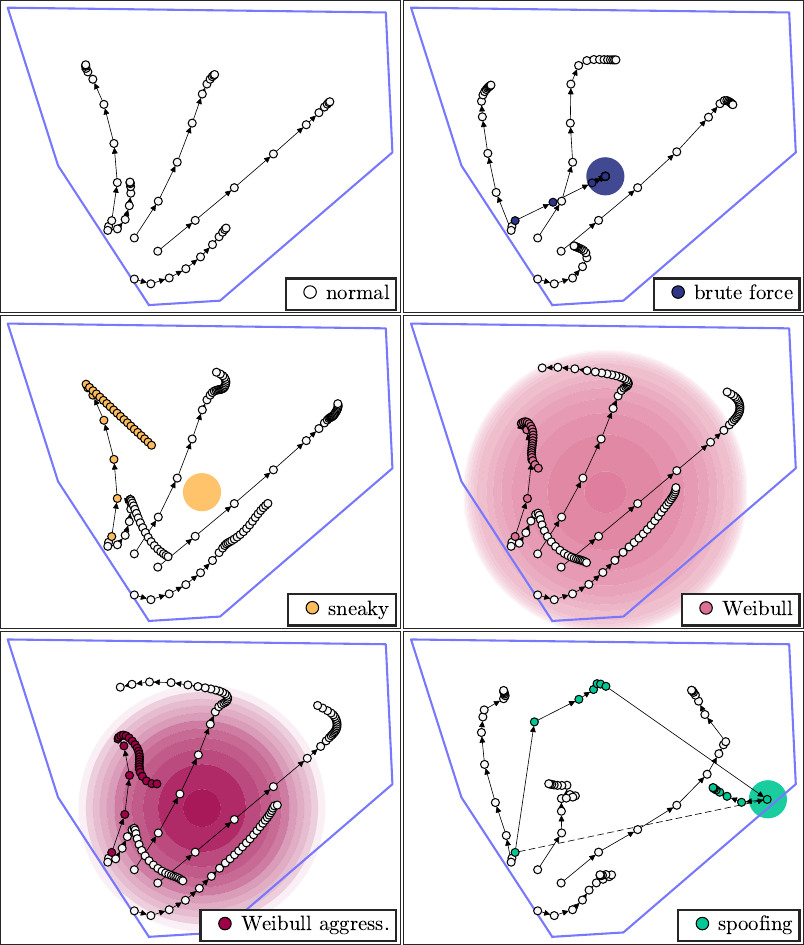}
		\caption[justification=left]{
			A deployment scenario visualizing normal swarm behavior and the behavior of the different antagonistic strategies. The antagonist's region of interest is either marked with a circle in the corresponding color or located at the center of the illustrated density. The true motion of the \textit{spoofing} robot toward its target is indicated by the dashed arrow.
		}
    \vspace*{-20pt}
		\label{fig:types}
	\end{center}
\end{figure}

Unfortunately, the robot swarm's endeavor to evenly distribute within the deployment area is susceptible to antagonistic attacks that exploit the robots' tendency to move away from each other.
Given a region of interest within the deployment area, a robot that moves towards this region automatically manipulates the Voronoi tessellation in a way that drives the other agents away from it.
As a result, an antagonistic agent may position itself strategically to assume control over the region of interest, thereby preventing the swarm's objective of detecting illicit activities or entities within that area.
Assuming that conspicuous deviations from normal behavior motions will facilitate the identification of antagonists, several types of antagonistic behavior are implemented and evaluated that differ in the way in which the antagonists approach the target area.
Figure~\ref{fig:types} depicts an example for each type of antagonist.
The first antagonistic type, named \textit{brute force} robot, simply moves in a straight line towards its antagonistic target by setting its target position $\bm{x}^*$ to the point $\bm{x}_{\text{ROI}} \in \mathbb{R}^2$ in the region of interest.
Thereby, it establishes a baseline for comparison with more advanced antagonists that attempt to reach their target in a way that conceals their antagonistic behavior.
The second, called \textit{sneaky}, antagonist refines the \textit{brute force} approach by exhibiting perfectly normal behavior at the beginning of the deployment task.
However, as soon as the \textit{sneaky} agent gets close to its optimal normal position, it starts to perform small motions towards its region of interest.
In this way, it imitates the step size of a robot that approaches its optimal target and avoids the large deviations from normal behavior motions exhibited by the \textit{brute force} agent.

A different approach of covertly moving towards its target region is taken by the \textit{Weibull} agent.
This antagonist modifies the normal behavior by weighting the area within its Voronoi cell according to its proximity to the region of interest.
With the goal of drawing antagonists that are distant from the region of interest toward $\bm{x}_{\text{ROI}}$, but transitioning to a more covert behavior when the agents approach the region of interest, a cumulative Weibull distribution $\phi: \mathbb{R} \rightarrow \mathbb{R}$ is used as a weighting function.
The robot's aggressiveness is tuned by modifying the shape $k  \in \mathbb{R}$, scale $\lambda  \in \mathbb{R}$, and shift $l  \in \mathbb{R}$ parameters of the distribution, yielding
\begin{equation}
	\begin{split}
		\bm{x}^* = \min_{\bm{x}} \left[ \int_{W} ||\bm{q} - \bm{x}||^2 \, \phi\left(-||\bm{q} - \bm{x}_{\text{ROI}}||^2 \right)\, \mathrm{d} W \right] \\
    \text{with }\, \phi(x, \lambda, k, l) = \begin{cases}
			1 - e^{-\left( (x - l)/ \lambda \right)^k} + 0.1 & \text{if } x \geq l \\
			0.1 & \text{else.}
			\end{cases}
			\label{eqn:eqn_vor_weighted}
	\end{split}
\end{equation}

The last, called \textit{spoofing}, antagonist moves straight towards the region of interest, but conceals its behavior by sending incorrect information about its position to the swarm until the target position $\bm{x}_{\text{ROI}}$ has been reached.
With $\bm{x} \in \mathbb{R}^2$ being the actual robot position and $\bm{x}_{\text{comm}} \in \mathbb{R}^2$ containing the position communicated to the swarm, the robot's behavior can be described by
\begin{equation}
	\begin{split}
	\bm{x}_{\text{comm}} &= 
		\begin{cases}
			\min_{\hat{\bm{x}}} \left[ \int_{W} ||\bm{q} - \hat{\bm{x}}||^2 \, \mathrm{d} W \right] & \text{if } \bm{x}_{\text{ROI}} \text{ not reached,} \\
			\bm{x} & \text{else,}
		\end{cases}\\
	\bm{x}^* &= 
		\begin{cases}
			\bm{x}_{\text{ROI}} & \text{if } \bm{x}_{\text{ROI}} \text{ not reached,} \\
			\min_{\hat{\bm{x}}} \left[ \int_{W} ||\bm{q} - \hat{\bm{x}}||^2 \, \mathrm{d} W \right] & \text{else.}
		\end{cases}
	\end{split}
	\label{eqn:eqn_spoofing}
\end{equation}

This results in the robot swarm perceiving a single anomalous action when the \textit{spoofing} agent switches from mimicking normal behavior to reaching the region of interest and communicating its correct position.
However, the spatial distance between the last mimicked action and the first correct action might not conform to the physical restrictions on robot motions.
Therefore, it potentially poses an out-of-distribution input to the anomaly detection method.

\section{Detection of anomalous actions based on their probability of occurrence}
\label{sec:approach}

Observing the robot motions in Figure~\ref{fig:types}, it becomes apparent that the categorization of a robot action as normal or anomalous is dependent on the robot's context, i.e., their position within the deployment area and relative to other swarm agents.
For instance, the normal robots modify their motions based on the antagonist's strategy while continuing to pursue their objective of optimizing the area's coverage.
As a consequence, a contextual anomaly detection approach is required in order to categorize the robot actions~\cite{ChandolaBanerjeeKumar09}.
Considerations regarding the suitable representation of this context are discussed in Section~\ref{sec:context}.

As previously explained, we must avoid restricting the approach to the availability and comprehensiveness of previous information on anomalies.
Thus, the proposed method intends to define a reference point for normal robot behavior~\cite{JeffreyTanVillar23} based on a dataset containing simulations or observations of normal robot behavior.
Statistical anomaly detection methods assume that such a dataset represents samples from an underlying distribution $p_{\bm{a}}^*$, where the normal behavior lies in regions with high probability density~\cite{ChandolaBanerjeeKumar09}.
Consequently, we can define the anomaly score of an observed action as the negative likelihood of a robot's action given the robot's current context.
In the present work, as detailed in Section~\ref{sec:nn}, the anomaly score is approximated by a neural network that has been trained on the dataset of normal robot behavior while using an unsupervised learning approach.

Eventually, in order to categorize a robot as normal or antagonistic, the estimated anomaly score is processed by a detection criterion, as described in Section~\ref{sec:criteria}.

\subsection{Finding a suitable representation for context and behavior}
\label{sec:context}

For the chosen implementation of normal coverage behavior described in Section~\ref{sec:coverage}, the context that determines the motion of a robot could potentially be reduced to a representation of the robot's Voronoi cell.
However, considering learned, or non-deterministic swarm motions, the detection method should not be restricted to this specific behavior.
Instead, we aim to find a more generally applicable context representation which requires little a priori knowledge about the algorithmic behavior of normal robots.

The context information available in the investigated coverage scenario comprises the shape of the deployment area and the robot's own position within the deployment area.
Additionally, the robots communicate their positions to other members of the robot swarm.
Based on this information, a robot's position relative to its environment is computed as the distance vectors between the robot and the positions of all other swarm agents as well as the distance vectors to number of points on the deployment area's borders.
These points include the corners of the deployment area, and the point on each edge between the area's corners that is closest to the robot.
Each distance vector is normalized and concatenated to a logarithmic scaling of the distance and to a two-dimensional label that indicates whether the distance corresponds to another robot or to a point on the border of the deployment area.
Since the number of context features $n_\text{s}$ in the resulting context $\bm{s}~\in~\mathbb{R}^{n_\text{s} \times 5}$ depends on the shape of the deployment area and the size of the robot swarm, it varies for each coverage task.
This complicates the representation of the context as a fixed-size input that can be processed by the neural network that predicts the robot behavior.
Thus, the context is preprocessed by passing it into a Long Short-Term Memory (LSTM) network~\cite{HochreiterSchmidhuber97}, which computes a fixed-size embedding of the context.
To leverage the irrelevance of the order of the spatial context features and to increase the network's capacity, a bidirectional LSTM is used~\cite{SchusterPaliwal97}.

In contrast, the method proposed in~\cite{WengerEbelEberhard24} uses one Long Short-Term Memory network to embed the distance vectors between the robots and a second LSTM to embed the distance vectors to the area vertices.
However, since the relevance of a vertex can be impacted by the presence of a neighboring robot close to the vertex and vice versa, the usage of separate networks might aggravate the recognition of such interdependencies.
The additional consideration of the area's edges is motivated by observations from~\cite{WengerEbelEberhard24}.
There, the behavior predictions indicated that the area's corners might not suffice to correctly assess the area's shape, especially for long edges between the corners.

On top of the modifications to the context representation, the present work proposes to represent the robot's motion $\bm{a} \in \mathbb{R}^3$ as a concatenation of the normalized motion vector and the motion's magnitude. 
This adaptation is motivated by the qualitative results in~\cite{WengerEbelEberhard24}, where artifacts in the prediction of the robots' behavior showed that the neural network had difficulties with the correct estimation of the direction and magnitude of the robot motion.

In summary, the novel key developments proposed by this work include the modified representation of the robot motions, the consideration of the area's edges, and the use of labels for the representation of the distance features that are passed into a single bidirectional LSTM.

Since the robot actions depend on the context, we assume that the detection of all types of antagonists benefit from an improved state embedding.
However, this change might concern the \textit{Weibull} and \textit{sneaky} behavior types in particular, since the small motion deviations performed by these types require a rather precise prediction of actions.

\subsection{Learning the likelihood of a robot action}
\label{sec:nn}

Given the current context $\bm{s}$ of the robot swarm distribution, the anomaly score of a robot action $\bm{a}$ is defined as its negative likelihood $- p_{\bm{a}}^* (\bm{a} | \bm{s})$ under the true data distribution $p_{\bm{a}}^*$ of normal robot behavior.
Since $p_{\bm{a}}^*$ is unknown, this paper uses a normalizing flow~\cite{TabakTurner13, RezendeMohamed15} to estimate the true likelihood value.
Normalizing flows are a type of neural network that has been proven to be powerful to model an unknown and possibly complex probability density $p_{\bm{a}}^*$.
In order to learn a suitable model $p_{\bm{a}}$ from the data, a normalizing flow with parameters $\bm{ \theta }$ is trained to maximize the probability density values 
$
	p_{\bm{a}}(\bm{a} \vert \bm{s}; \bm{ \theta })
$ of the actions in the training data.
This training data exclusively contains simulations of normal, i.e., functional and cooperating robot behavior.
Details on the architecture and training of the normalizing flow are provided in the following.

The network learns an invertible and differentiable transformation function $\bm{T} = \bm{T}_K \circ ... \circ \bm{T}_1$ between $p_{\bm{a}}^*$ and a simple base distribution $p_{\bm{u}}$, e.g., a uniform distribution.
As explained in~\cite{PapamakariosEtAl21}, normalizing flows compute a probability density value
\begin{equation}
	p_{{\bm{a}}}( \bm{a} | \bm{s}; \bm{\theta}) = p_{{\bm{u}}}\!\left(\bm{T}^{-1}(\bm{a} | \bm{s}; \bm{\theta})\!\right)\, \vert\!\det \!\left(\bm{J}_{\bm{T}^{-1}}(\bm{a})\!\right)\!\vert
	\label{eqn:p_NF}
\end{equation}
with $\bm{J}_{\bm{T}^{-1}}$ being the Jacobian matrix of the inverse transformation function $\bm{T}^{-1}$ with respect to $\bm{a}$, thus performing a change of variable.
Additionally, we can easily sample from $p_{\bm{a}}$ by computing
$
	\bm{a}_\text{s} = \bm{T}(\bm{u}_\text{s})
$
for a sample $\bm{u}_\text{s} \sim p_{{\bm{u}}}$~\cite{PapamakariosEtAl21}.
During training, the normalizing flow approximates the true data distribution $p_{\bm{a}}^*$ by maximizing the probability of training samples $\bm{a} \sim p_{\bm{a}}^*$ under the model distribution $p_{\bm{a}}$.
This is equivalent to minimizing the Kullback-Leibler divergence
\begin{align}
	\mathcal{L}(\bm{\theta}) &= D_{\text{KL}}[\,p^*_{\bm{a}}(\bm{a} | \bm{s})\,\vert \vert\,p_{{\bm{a}}}(\bm{a} | \bm{s}; \bm{\theta})\,] \nonumber
	= -\mathbb{E}_{p^*_{\bm{a}}(\bm{a} | \bm{s})} [\, \log p_{{\bm{a}}}(\bm{a} | \bm{s}; \bm{\theta}) \,] + \text{const.} \nonumber\\
	&\approx - \frac{1}{N} \sum_{n=1}^{N} \log p_{{\bm{u}}} \!\left(\bm{T}^{-1}(\bm{a}_n | \bm{s}; \bm{\theta})\!\right) + \log \vert\! \det \!\left(J_{\bm{T}^{-1}}(\bm{a}_n | \bm{s}; \bm{\theta})\!\right)\!\vert + \text{const.} \label{eqn:loss}
  \end{align}
between $p_{\bm{a}}^*$ and $p_{\bm{a}}$~\cite{PapamakariosEtAl21}.
Both in the present work and in our previous work~\cite{WengerEbelEberhard24}, each transformation $\bm{T}$ is computed using a strictly monotonic spline-based transformer function with $S$ spline segments that are modified based on the current context information $\bm{s}$. 
While the present work investigates the use of a non-autoregressive coupling flow, the network used in~\cite{WengerEbelEberhard24} is a masked autoregressive flow with both $\bm{s}$ and $\bm{a}$ influencing the transformation.
For details please refer to~\cite[Sec.\ 3.2]{WengerEbelEberhard24}.
The normalizing flows are built based on the \texttt{Python} package \texttt{nflows}~\cite{nflows}.

\subsection{Criteria for categorizing an agent as antagonistic}
\label{sec:criteria}

Given the approximated likelihood $p_{\bm{a}} (\bm{a} | \bm{s}; \bm{ \theta })$ as a measure of normalcy for $\bm{a}$, a threshold $h \in \mathbb{R}$ for normal behavior is defined, such that the fulfillment of the condition
\begin{equation}
	p_{\bm{a}}(\bm{a} \vert \bm{s}; \bm{ \theta }) \geq h
	\label{eqn:a1}
\end{equation}
categorizes an action $\bm{a}$ as normal.
An increase of this threshold $h$ therefore leads to more actions being categorized as anomalous.
This increases the number of anomalous actions that are recognized correctly, but also the false positive rate, i.e., the percentage of normal actions that are incorrectly categorized as anomalous.
Since the tolerable percentage of incorrectly categorized agents depends on the task, the threshold $h$ can be adapted according to the desired maximum false positive rate $FPR_{\max}$ of normal actions $\bm{a_{\text{normal}}}$, such that
\begin{equation}
	P\left(~p_{\bm{a}}(\bm{a_{\text{normal}}} \vert \bm{s}; \bm{ \theta }) < h~\right) < FPR_{\max}.
\end{equation}
Naively, an agent can be categorized as anomalous if it once performs an action that does not fulfill Equation~\ref{eqn:a1}.
However, with robot agents performing a sequence of actions during a deployment task, the probability of a robot to be labeled as anomalous increases with the number of performed actions.
Thus, while this naive approach might be successful in detecting antagonists, it is expected to yield a large false positive rate for normal agents which is higly undesirable.

The probability of observing a number $k$ of anomalous actions in a sequence of observed actions of length $|\bm{a}_{\text{o}}|$ can be determined by using the binomial distribution.
We hence modify the detection criterion to categorize an agent as anomalous if
\begin{equation}
	\begin{split}
	&\left[ \binom{|\bm{a}_{\text{o}}|}{k}f_\text{p}^k(1-f_\text{p})^{|\bm{a}_{\text{o}}|-k} \right] < FPR_{\max} \\
	& \text{ with }  ~k := \sum_{\bm{a} \in {\bm{a}}_{\text{o}}} \left[~p_{\bm{a}}(\bm{a} \vert \bm{s}; \bm{ \theta }) < h~\right].
	\end{split}
	\label{eqn:a_binom}
\end{equation}
However, a potential drawback of the criterion is the loss of information about the exact probability density value $p_{\bm{a}}$.
In fact, all actions whose prediction $p_{\bm{a}}$ falls below the threshold $h$ are equally likely to cause an agent to be labeled as anomalous.

To alleviate this issue, a third detection criterion is introduced.
It calculates the mean over the predicted log probability values of all actions $\bm{a}_{\text{o}}$ that have been performed by a robot during the current deployment task, i.e.,
\begin{equation}
	\left[ \frac{1}{|\bm{a}_{\text{o}}|} \sum_{\bm{a} \in {\bm{a}}_{\text{o}}} \log p_{\bm{a}}(\bm{a} \vert \bm{s}; \bm{ \theta }) \right] < h.
	\label{eqn:a_mean}
\end{equation}
Since the logarithmic function causes a disproportional impact of very small probability density values, a normal agent is required to actuate close to the expected behavior.
Additionally, under the assumption of independent and identically distributed action predictions and according to the central limit theorem~\cite{Polya20}, the mean log probability of normal agents should approximate a normal distribution.
Consequently, a robot that repeatedly performs actions that are only slightly larger than $h$ and thus is considered normal according to Equation~\eqref{eqn:a_binom} might still be correctly labeled as anomalous based on Equation~\eqref{eqn:a_mean}.

\section{Setup of the deployment scenario for simulations and hardware experiments}
\label{sec:experiments}

The following setup is used both for simulations and for the hardware experiments with Table~\ref{tab:data} displaying more detailed information on relevant parameter values.
\begin{table*}[htb]
	\caption[]{Parameters used for the collection of simulation and hardware data.}
	\centering
	\begin{tabular}{lcrcr}
	\toprule
	{parameter} & \hspace*{15pt} & {simulation} & \hspace*{15pt} & {hardware} \\ 
	\cmidrule(rl){1-5}
	swarm robot agents $n_\text{r}$ && 3 {--} 20 & &3 {--} 5\\
	area size per agent && 1 {--} 25 m$^2$ && 0.6 {--} 3.6 m$^2$\\
	deployment area vertices &&  3 {--} 8 && 3 {--} 6\\
	$x$ / $y$ ratio of area & & 0.2 {--} 5 & &0.7 {--} 3.2\\
    maximum steps $t_{\max}$ && 50 & &15\\
    convergence threshold && 7.5 cm && 5 cm\\
    communication interval $\Delta t_\text{c}$ && 3 s & &3 s\\
    control interval $\Delta t_\text{v}$ && 0.2 s & &0.2 s\\
	\bottomrule
	\end{tabular}
	\label{tab:data}
\end{table*}

Initially, an arbitrary deployment area is defined according to the investigated situation.
Depending on the area size, a suitable number $n_\text{r}$ of robot agents is selected to build a robot swarm.
The robots distribute from their initial positions within the deployment area by iteratively optimizing their position depending on the positions of the other agents, as described in Lloyd's~Algorithm~\cite{Lloyd82}.
Each iteration starts with the robots communicating their current position to all swarm members.
Based on the positions received from the other agents as well as the outline of the deployment area, a robot computes its Voronoi cell.
For the remainder of the time interval $\Delta t_\text{c}$ until the next communication with the swarm, each robot optimizes the coverage of its Voronoi cell while revising and adapting its target velocity in regular time intervals $\Delta t_\text{v}$.
The deployment task ends when the swarm robots have converged towards their final position and stop moving or after a predefined maximum number of steps $t_{\max}$.
Thus, the data available for detecting an anomalous agent is restricted to the information communicated between agents.
This information includes the agents' positions $[\bm{x}_t, \bm{y}_t] \in \mathbb{R}^{n_\text{r} \times 2}$ and the resulting actions $[\bm{a}_x, \bm{a}_y] \coloneqq \begin{bmatrix}
	\bm{x}_{t + \Delta t_\text{c}} - \bm{x}_t &
	\bm{y}_{t + \Delta t_\text{c}} - \bm{y}_t
\end{bmatrix} \in \mathbb{R}^{n_\text{r} \times 2}$.

\subsection{Preprocessing the state and action data}
\label{sec:prep}

The action $\bm{a}  = \begin{bmatrix}a_x & a_y \end{bmatrix}^\top$ of a robot can either serve directly as an input to the normalizing flow, or can be represented by its normalized direction and its magnitude relative to the maximum admissible magnitude $a_{\max}$, i.e.,
$
 	\bm{a}_\text{norm} = \begin{bmatrix}
		a_x /\left\lVert \bm{a}\right\rVert &
		a_y /\left\lVert \bm{a}\right\rVert &
		\left\lVert \bm{a}\right\rVert / a_{\max}
	\end{bmatrix}^\top
$.

To represent the position of a robot relative to the robot swarm and the deployment area, the distance vectors $\bm{d} \in \mathbb{R}^{2}$ between each robot position and a set of environmental entities are computed.
These entities include the other robots' positions, the vertices of the deployment area and the perpendicular foot of the robot's distance vector to the edges of the area.
Entities that are in the proximity of a robot are generally more relevant for the robot's motions.
Therefore, we choose to nonlinearly scale the distance, resulting in a distance feature
$
	\bm{d_\text{norm}} = \begin{bmatrix}
		d_x / \left\lVert \bm{d}\right\rVert &
		d_y / \left\lVert \bm{d}\right\rVert & 
		\log(\left\lVert \bm{d}\right\rVert) / \bm{d}_{\max}
	\end{bmatrix}
$,
 that again separately encodes the information on direction and magnitude.
To build the context $\bm{s}$, a $'10'$ label is concatenated to each distance features between a robot and the border of the deployment area and a $'01'$ label is concatenated to each distance to another robot agent. 
Additionally, the distance features within $\bm{s}$ are randomly shuffled. 

\subsection{Hardware and multibody model of the robot agents}
\label{sec:hardware}

\begin{figure}[htb]
	\begin{center}
		\includegraphics[width=0.47\textwidth]{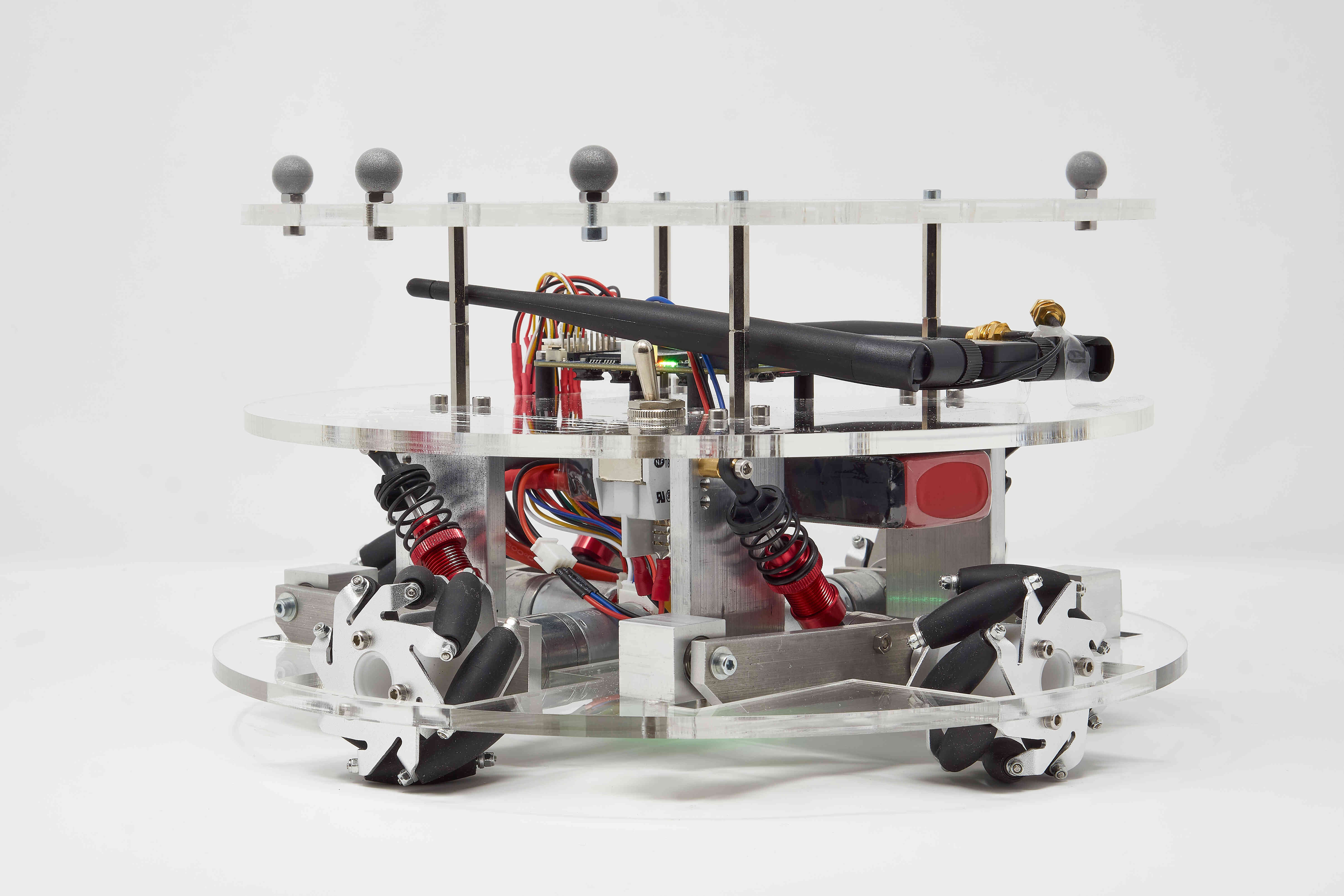}
		\includegraphics[width=0.45\textwidth]{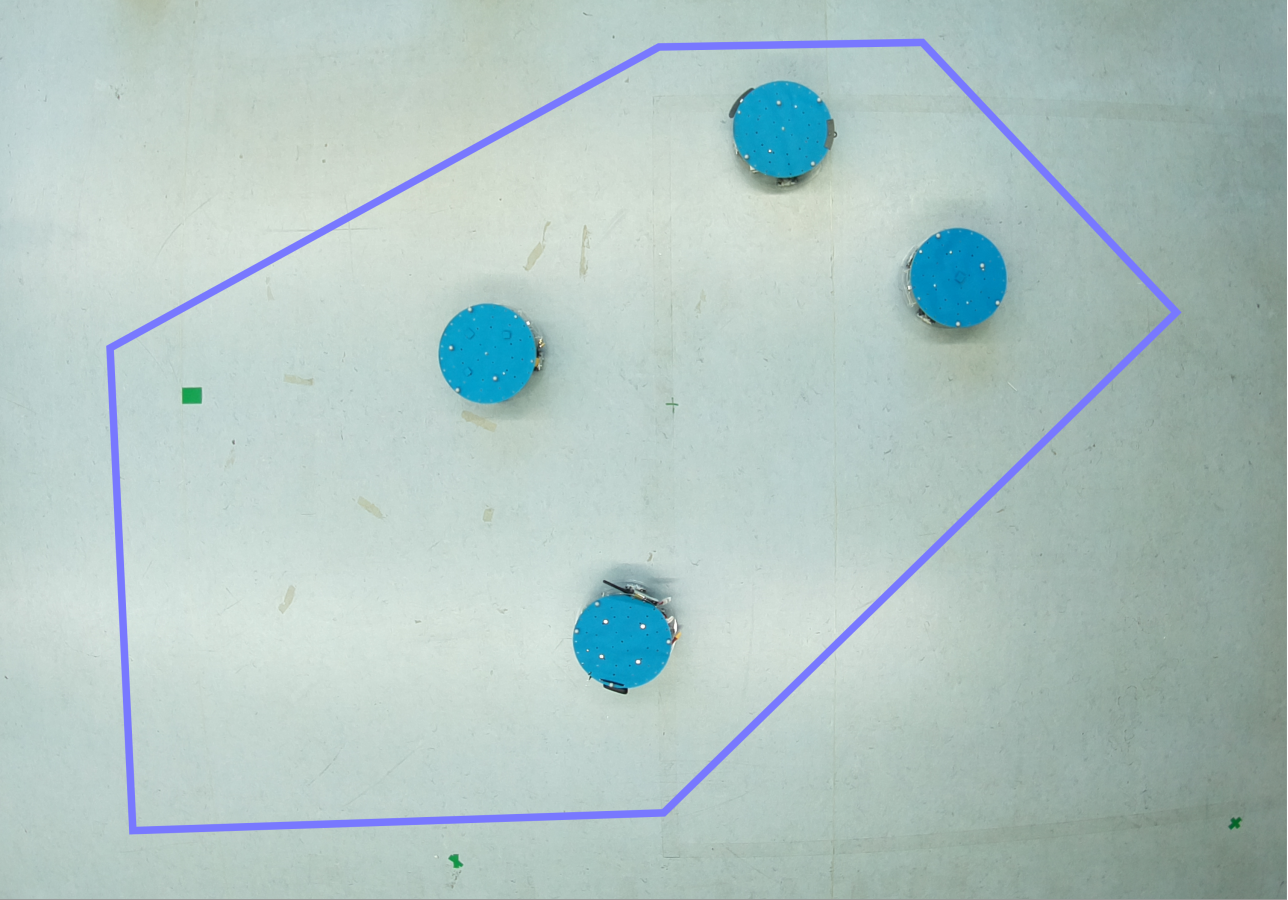}
		\caption[justification=left]{
		Omnidirectional mobile robot (\textit{left}) used in the hardware robot swarm (\textit{right})~\cite{Ebel21}. The blue lines indicate the deployment area.}
		\label{fig:robot}
		\vspace*{-25pt}
	\end{center}
\end{figure}

Figure~\ref{fig:robot} depicts the build type of the robots employed in this work, an omnidirectional mobile robot with four Mecanum wheels, called Holonomic Extensible Robotic Agent (HERA) \cite{EbelEberhard21}.
The robots are controlled by passing the desired translational velocity of the robot's center and calculating the corresponding angular velocities for the wheels
and the wheel motions are realized by local motor controllers running at a frequency of $100\,\textnormal{Hz}$.
The robots can reach a maximum velocity of $0.4\, \text{m}/\text{s}$ in each direction.
Their positions are tracked with a frequency of $100\,\textnormal{Hz}$ using an Optitrack camera system with six Optitrack Prime 13W cameras and the swarm communicates via the message passing system LCM \cite{HuangOlsonMoore10}.
For simulations, we use the multibody model of the hardware robot that is derived in~\cite{Ebel21}.

\subsection{Training and tuning of the baseline and the proposed neural network}

The best neural network configuration found in~\cite{WengerEbelEberhard24} is used as a baseline for comparisons.
To allow a direct comparison of their validation performance, the baseline network is retrained on the training dataset until convergence or ceasing to improve.
For training, a Gainward GeForce RTX 4090 Phantom GS with 24 GB VRAM is used.
The training data contains 350 randomly generated deployment tasks, resulting in 37\,249 state-action samples, while the training is validated on 125 simulation runs.
The baseline network has 19\,212 trainable weights and is trained for 376 epochs with a final validation log probability of $ \log p_{{\bm{a}}}(\bm{a_{\text{val}}} | \bm{s}_{\text{val}}; \bm{\theta}) = 1.957$, while the best network found with the new configuration has 25\,887 trainable weights and trained for 647 epochs with $ \log p_{{\bm{a}}}(\bm{a_{\text{val}}} | \bm{s}_{\text{val}}; \bm{\theta}) = 5.765$. Both neural network are implemented in \texttt{PyTorch}~\cite{PaszkeEtAl19} and their hyperparameters were optimized via the Bayesian hyperparameter optimization provided by Weights and Biases \texttt{(WandB)}~\cite{wandb}. 

As introduced in Section~\ref{sec:criteria}, the thresholds $h$ for the detection criteria were tuned to a false positive rate of $FPR_{\max} = 5\%$ and $FPR_{\max} = 1\%$ on 100 additional simulations of normal swarm behavior.
The simulated test dataset contains 1000 deployment runs per antagonistic type, whereas the hardware data is collected on 206 deployment runs with at least 29 runs per antagonistic strategy.
Each scenario in the test dataset includes a single antagonistic agent.

\section{Results}
\label{sec:result}

To evaluate the performance of the proposed method, several metrics are computed on the test dataset.
The sensitivity or true positive rate (TPR) refers to the ratio of agents that are correctly categorized as antagonists ($\hat{\text{A}} | \text{A}$) to the total number of agents that show antagonistic behavior (A).
Similarly, the specificity or true negative rate (TNR) describes the number of agents that are correctly categorized as normal by the detection algorithm ($\hat{\text{N}} | \text{N}$) over the total number of agents that exhibit normal behavior (N).
Thus, sensitivity and specificity indicate how accurate the prediction of the normalizing flow is for the different types of robot agents.
Additionally, the precision or positive predictive value (PPV) describes the ratio of antagonists that were correctly identified as anomalous ($\hat{\text{A}} | \text{A}$) to all robots categorized as anomalous ($\hat{\text{A}}$), such that
$$ 
\text{TPR} = \frac{\hat{\text{A}} | \text{A}}{\text{A}}~,\hspace{10pt} 
\text{TNR} = 1 - \text{FPR} = \frac{\hat{\text{N}} | \text{N}}{\text{N}}~,\hspace{10pt} 
\text{PPV} = \frac{\hat{\text{A}} | \text{A}}{\hat{\text{A}}} = \frac{\hat{\text{A}} | \text{A}}{\hat{\text{A}} | \text{A} + \hat{\text{A}} | \text{N}}. $$

\subsection{Simulation results}

\begin{figure}[hbtp]
	\begin{center}
		\includegraphics[width=\textwidth]{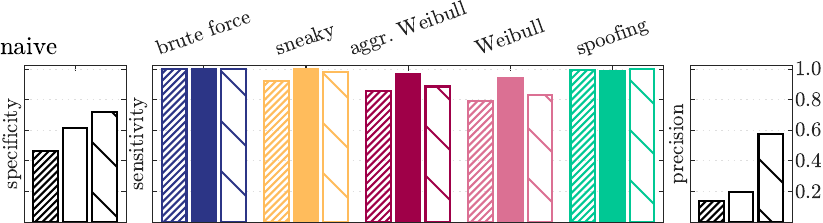} \\
		\vspace*{4pt}
		\includegraphics[width=\textwidth]{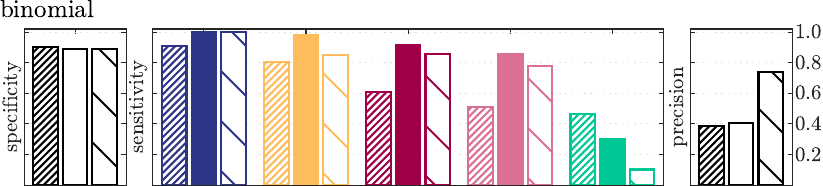} \\
		\vspace*{4pt}
		\includegraphics[width=\textwidth]{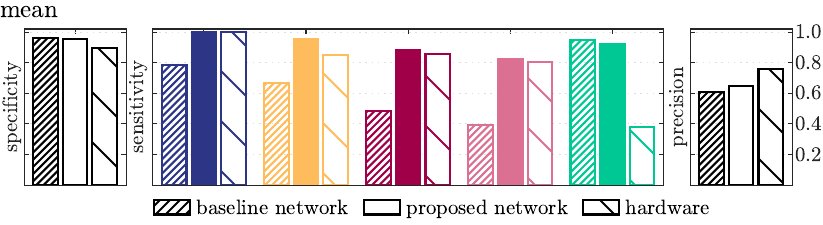}
		\vspace*{-20pt}
		\caption[justification=left]{
			Specificity, sensitivity and precision results for the baseline and the proposed network on the naive detection criterion, binomial detection criterion, and mean detection criterion. Higher values indicate a better performance. The allowed maximal false positive rate of normal agents is set to 5\%.
		}
		\label{fig:acc}
		\vspace*{-15pt}
	\end{center}
\end{figure}

Figure~\ref{fig:acc} illustrates the performance of the proposed neural network with the considered baseline network~\cite{WengerEbelEberhard24} on the three detection criteria introduced in Section~\ref{sec:criteria}.
With larger values representing a better performance, the proposed network (filled bars) clearly outperforms the baseline (striped bars) for almost every combination of detection criterion, robot type, and performance metric.
This difference in performance increases for more sophisticated and covert antagonistic behavior, as exemplified by the \textit{Weibull} agent.
The sensitivity results for the \textit{brute force}, \textit{sneaky}, \textit{aggressive Weibull} and \textit{Weibull} agents indicate a positive correlation between the aggressiveness of the antagonistic behavior and the corresponding detection accuracy.
These results emphasize the importance of the trained network being able to precisely predict the robot motions, thereby allowing to identify small deviations from the expected behavior.

As expected, the naive categorization described by Equation~\ref{eqn:a1} is sensitive to antagonistic agents, but leads to a higher false positive rate, i.e., a low specificity, for normal agents due to not considering the number of actions performed during a deployment run.
Thus, this criterion is not considered further.
The binomial and mean detection criteria show an improved specificity for the normal robot agents.
Nevertheless, only the mean criterion adheres to the desired maximum false positive rate of $5\%$.
The most obvious difference between the binomial and mean detection criteria concerns the sensitivity on the \textit{spoofing} robot.
While the binomial criterion seems to be slightly more robust to the multiple anomalous actions performed by the other anomalous types, it is unsuitable for detecting the single anomalous action communicated by the \textit{spoofing} agent.
In contrast, the unlikely and possibly out-of-distribution action has a significant impact on the computation of the mean in Equation~\ref{eqn:a_mean}.

\begin{figure}[hbtp]
	\begin{center}
		\includegraphics[width=\textwidth]{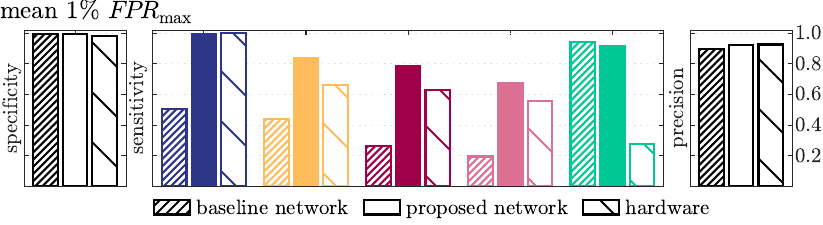}
		\vspace*{-20pt}
		\caption[justification=left]{
			Results for the mean criterion using an allowed maximal $FPR_{\max}$ of 1\%.
		}
		\label{fig:acc1}
		\vspace*{-25pt}
	\end{center}
\end{figure}

With the number of normal agents being more than 50 times higher than the number of antagonistic agents, the precision value is heavily influenced by the number of false positives $\hat{\text{A}} | \text{N}$.
Consequently, the high specificity, but also the good sensitivity values of the mean criterion result in the highest precision value compared to the other methods.
Overall, the combination of the proposed neural network and the mean detection criterion yields the most reliable, consistent performance.
As shown in Figure~\ref{fig:acc1} the mean criterion's $FPR_\text{max}$ is easily tunable to different tasks, taking the size of the robot swarm and the prioritization of false negatives into account.

\subsection{Hardware results}

\begin{figure}[hbtp]
	\begin{center}
		\includegraphics[width=\textwidth]{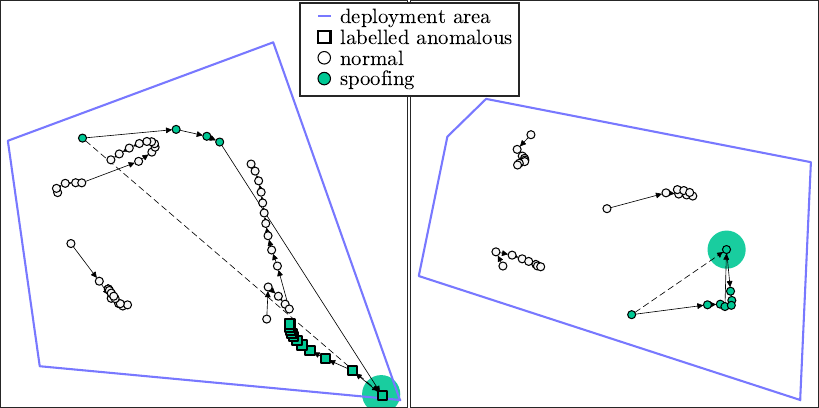}
		\caption[justification=left]{
			Two different hardware runs containing a spoofing agent. While the antagonist in the left scenario performs a large spoofing action and is correctly categorized, as indicated by the square markers, the small spoofing action of the antagonist on the right is not recognized by the mean detection method.
		}
		\vspace*{-15pt}
		\label{fig:spoofing}
	\end{center}
\end{figure}

The deployment runs executed in hardware are evaluated for each of the three detection criteria.
As described in Section~\ref{sec:experiments}, the network has only been trained on simulated data and the conditions of the hardware experiments slightly differ from the simulated scenario due to spatial restrictions.
Possible issues like communication interference or environmental noise are not accounted for in the training data.
Nevertheless, the proposed method yields good results on the hardware data, as visualized by the right-hand bars in Figure~\ref{fig:acc} with slight decreases in the sensitivity for all detection methods.
This performance difference between simulation and hardware is amplified when requiring small false positive rates, as indicated in Figure~\ref{fig:acc1}.
A notable difference can be observed for the spoofing strategy.
Since only a small deployment area is available for the hardware experiments, the magnitude of the spoofing motion is smaller than in the simulations.
This can cause the binomial and mean detection methods to incorrectly categorize the spoofing antagonist, as shown in the exemplary hardware runs in Figure~\ref{fig:spoofing}.

In contrast, the naive detection method has a very high sensitivity for the spoofing strategy, since this method only requires a single anomalous action to categorize an agent as antagonistic.
Additionally, the smaller deployment area causes the robots to execute fewer actions, which benefits the specificity and precision of the naive approach.

\subsection{Qualitative results}
\label{sec:qual}

\begin{figure}[hptb]
	\begin{center}
		\includegraphics[width=\textwidth]{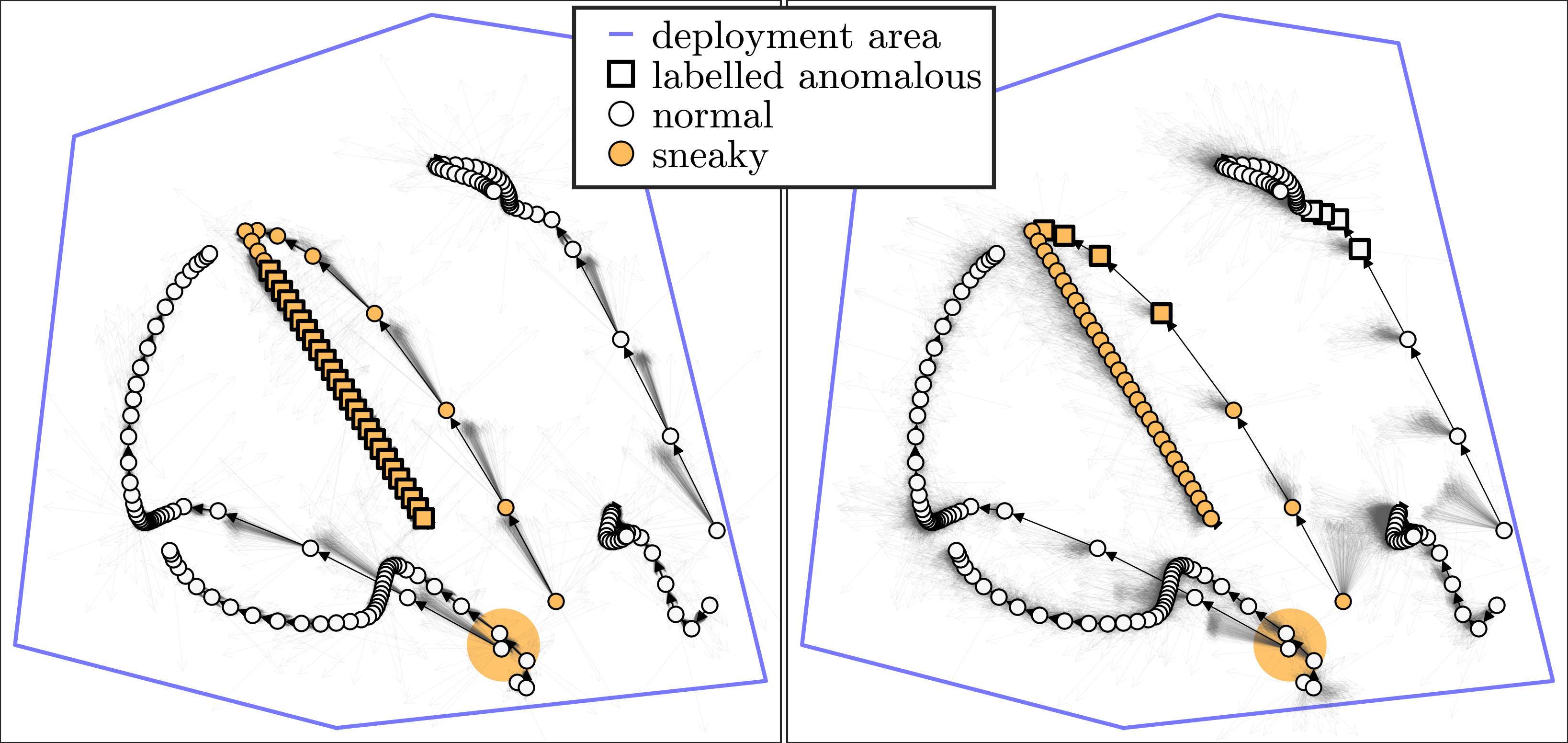}
		\includegraphics[width=\textwidth]{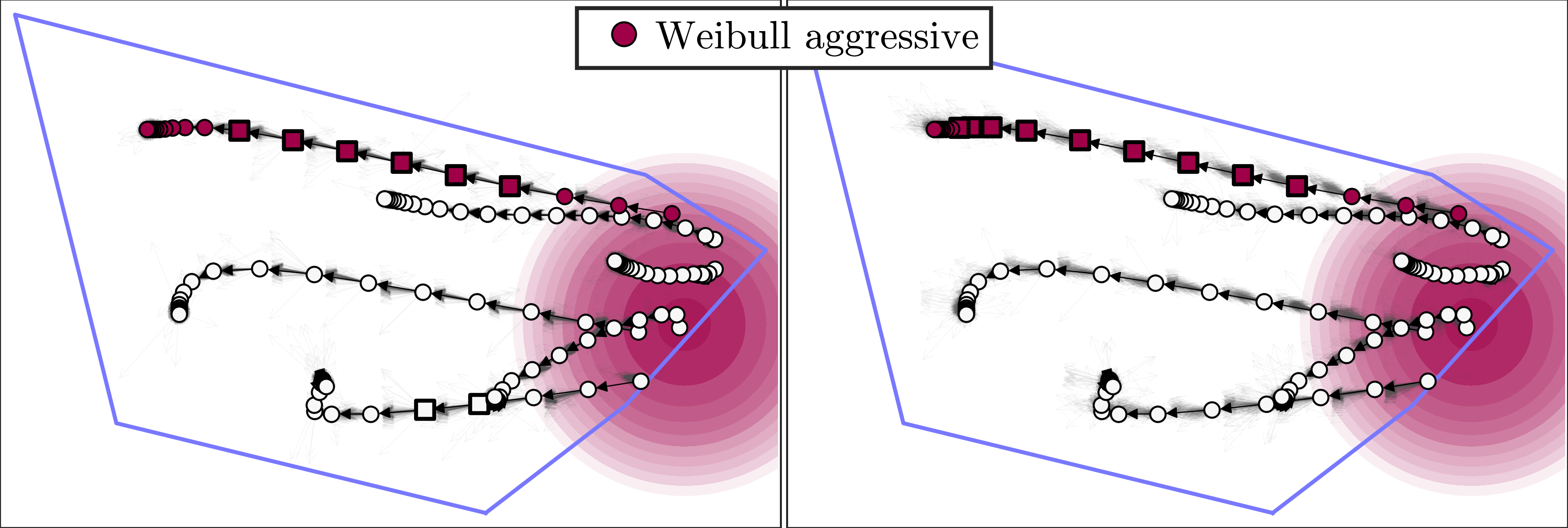}
		\includegraphics[width=\textwidth]{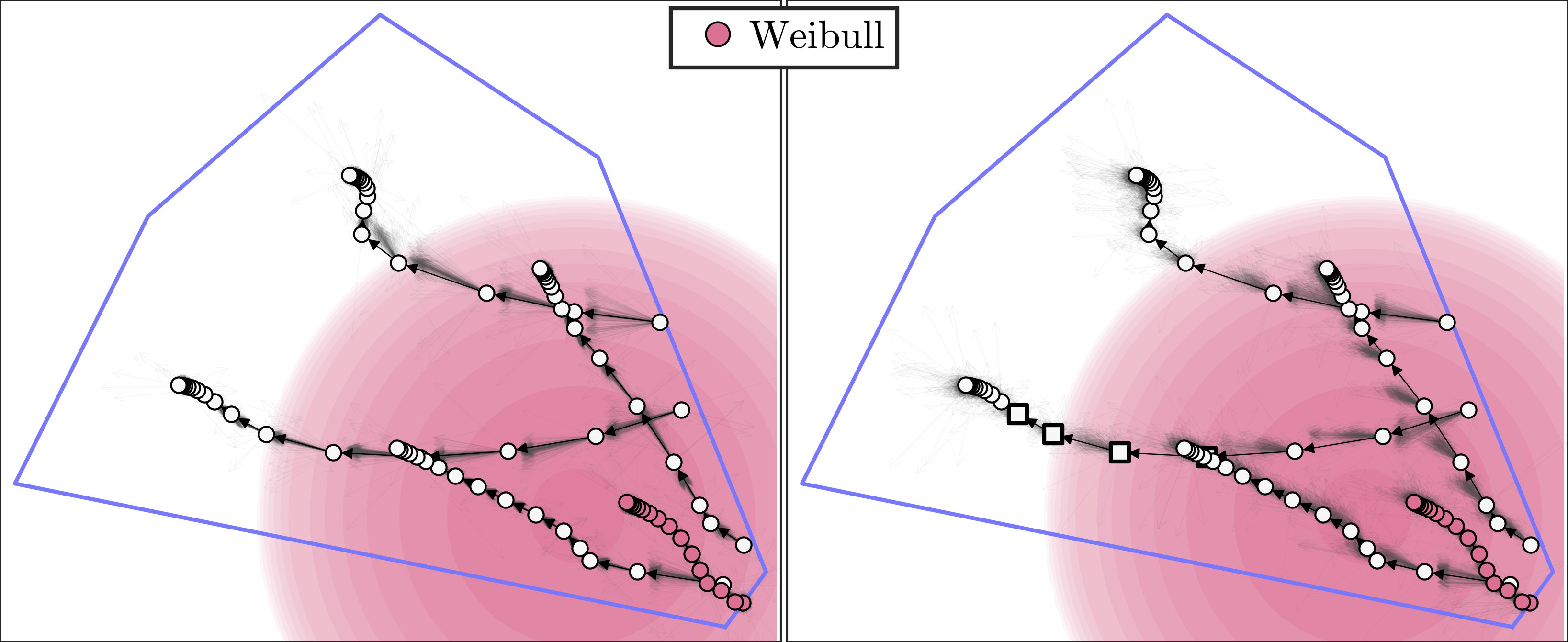}
		\caption[justification=left]{The rows visualize the application of the mean detection approach in three exemplary deployment scenarios. 
		Each row shows both the proposed network (\textit{left}) and the baseline (\textit{right}). The gray arrows approximate the action distribution predicted for each action. If an agent is drawn with a square, it has been labeled as anomalous.}
		\label{fig:qual}
		\vspace*{-14pt}
	\end{center}
\end{figure}

An impression of the prediction performance when using the mean detection criterion is given in Figure~\ref{fig:qual}, both for the neural network proposed in this work (left) and for the baseline network (right).

A black arrow represents the motion action performed by an agent during one communication interval $\Delta t_\text{c}$. 
After each $\Delta t_\text{c}$, the detection method updates its categorization of the agents as normal or anomalous and robot agents that are labelled as anomalous are marked with a square.
The action distribution $p_{\bm{a}}(\bm{a} \vert \bm{s}; \bm{ \theta })$ predicted by the normalizing flow is visualized by sampling actions $\bm{a} \sim p_{\bm{a}}(\bm{a} \vert \bm{s}; \bm{ \theta })$ for each robot and context $\bm{s}$ and plotting the samples as gray arrows.
Since the normalizing flows are trained to maximize the likelihood of normal actions, this action distribution should closely match the observed normal actions, but diverge from the actions of robots that show antagonistic behavior.
When comparing both networks, it is evident that the proposed network's prediction matches the observed normal actions quite well, while the baseline tends to compute a skewed or heavy-tailed distribution for normal actions.
These imprecise predictions lead to difficulties in detecting the \textit{sneaky} agent in the first scenario.

\newpage
One difficulty in detecting antagonists with the mean detection method is visualized by the \textit{aggressive Weibull} agent in the second scenario.
The agent is correctly identified as anomalous at the start of the task, but subsequently performs a number of normal actions that move the mean log probability above the threshold $h$ that classifies an agent as anomalous.
Whereas this behavior exploits a vulnerability of the mean detection method, it might also result in the robot being unsuccessful in reaching its antagonistic target, e.g., due to being pushed outside the influence of its corresponding Weibull distribution.
In fact, the sensitivity of detecting the \textit{aggressive Weibull} agent jumps from $0.883$ to $0.961$ when excluding unsuccessful antagonistic agents from the evaluation.
A second issue can be seen in case of the \textit{Weibull} agent in the third scenario, where, due to being favorably positioned at the start of the task, the agent is able to perform motions that are similar to the normal behavior while maintaining its control over the antagonistic target.

\begin{figure}[hbt]
	\vspace*{5pt}
	\begin{center}
		\includegraphics[width=\textwidth]{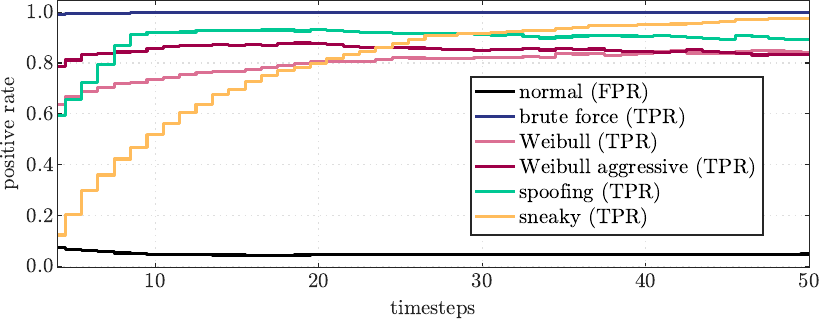}
	\end{center}
	\vspace*{-20pt}
	\caption[justification=left]{
		The true positive rate for antagonistic agents, and false positive rate for normal agents per time step of the deployment task. The agents are categorized using the mean criterion.
	}
	\label{fig:time}
\end{figure}

Considering that the classification of an agent as normal or antagonistic can change over the duration of the coverage task, we compute the antagonists' true positive rate per time step, as well as the normal agents' false positive rate, and visualize it in Figure~\ref{fig:time}.

The values of the positive rate over time are consistent with the observations from the qualitative results.
Except for the sneaky agent, which requires the most time to initiate its antagonistic behavior, the agents' positive rate stays remains relatively stable after 10 to 20 time steps.
The slight decrease likely corresponds to the aforementioned effect of agents that switch to normal behavior.

\subsection{Importance of the proposed methodological design choices}

\begin{figure}[htbp]
	\begin{center}
		\includegraphics[width=\textwidth]{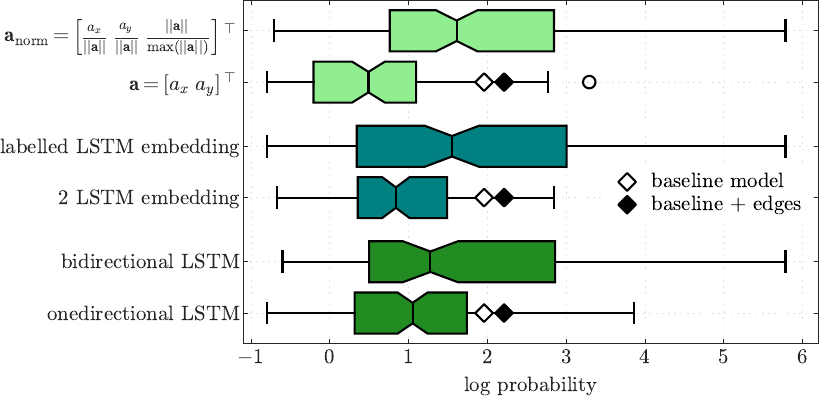}
		\vspace*{-20pt}
		\caption[justification=left]{
			The box plots visualize the effect of different hyperparameter configurations on the mean validation log probability of 250 trained neural networks while leaving aside all other hyperparameters. A larger log probability indicates a better performance. 
			The new hyperparameter settings clearly outperform the baseline hyperparameters.
		}
		\label{fig:configs}
		\vspace*{-10pt}
	\end{center}
\end{figure}

The approach proposed in this paper generally leads to a satisfying performance.
To better understand which methodological design choices are crucial for achieving the observed performance, an ablation study is conducted.
To that end, 250 neural networks with different hyperparameters were trained and their performance on the validation data was recorded.
Figure~\ref{fig:configs} visualizes the effect of the adaptations proposed in Section~\ref{sec:context} on the validation log probability.

Each pair of box plots only differs in the hyperparameter setting specified on the left while leaving aside the effect of all other parameters.
Additionally, the plot shows the validation log probability of the baseline network and the performance of the same baseline when passing not only the distance vectors to the vertices of the deployment area, but also the distances to the edges.
With a higher log probability indicating a better fit to the target distribution $p_{\bm{a}}^*$, the results suggest that the depicted adaptations lead to a strong improvement in performance.
This is further confirmed by computing a two-sided Brunner-Munzel test \cite{BrunnerMunzel00} on the validation log probability of neural networks trained with two-dimensional actions and three-dimensional actions ($W_N^{BF} = 2.58$, $p = 0.011$), labelled embedding and 2-LSTM embedding ($W_N^{BF} = 3.94$, $p < 0.001$), and bidirectional or one-directional LSTM ($W_N^{BF} = 9.58$, $p < 0.001$).

No obvious performance difference is found between the use of a masked autoregressive spline flow or a non-regressive coupling spline flow. 
Hence, for conciseness, that distinction is not depicted.

\subsection{Countermeasures against an antagonist}

\begin{figure}[htbp]
	\begin{center}
		\includegraphics[width=\textwidth]{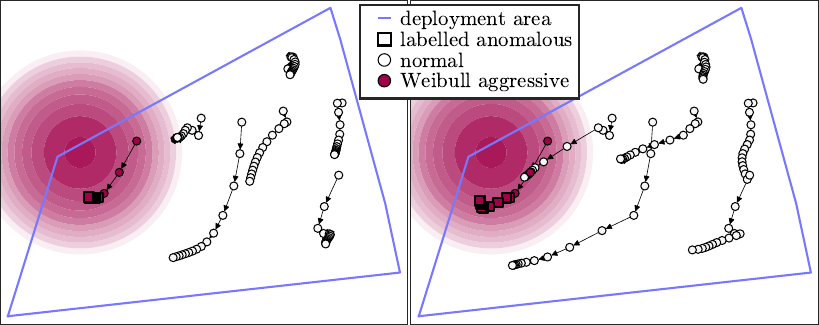}
		\caption[justification=left]{
			The figure on the right visualizes the effect of the robot swarm's immediate exclusion of the antagonist from the swarm once it is detected. This exclusion implies that the swarm covers the area without considering the antagonist's position. Consequently, one of the normal agents approaches the antagonist's region of interest and prevents the antagonist from obtaining exclusive control over its target.	
		}
		\label{fig:exclude}
		\vspace*{-15pt}
	\end{center}
\end{figure}

Given that an antagonist has been identified, an interesting research question is what countermeasures the swarm can take against it.
One simple strategy for the coverage scenario is the exclusion of the antagonist from the swarm.
This means that the swarm only takes information into account that has been sent by swarm members that are categorized as normal.
The effect of this exclusion is visualized in Figure~\ref{fig:exclude}.

It can be observed that the remaining robots are able to cover the area evenly, including the antagonistic target, and complete the deployment task with one agent less.
Thus, if a suitable detection method is available, already a simple strategy against antagonistic behavior might effectively prevent the antagonist from controlling a target region.
More advanced strategies will be investigated in future work.


\section{Conclusion}
\label{sec:conclusion}

The proposed approach exceeds the current state of the art for data-based contextual anomaly detection of physically antagonistic behavior performed by agents in a robot swarm.
As an ablation study has highlighted, due to new design choices for the representation of context and robot behavior and adaptations of the neural network architecture, the performance in predicting the probability density value of an action is significantly improved compared to previous methods.
Combined with the proposed mean criterion, the anomaly detection method is found to be both sensitive to deviations from the expected behavior of an agent and robust to the number of actions that the swarm executes.
Thus, a sensitivity of more than $80\%$ is achieved for all types of antagonistic agents that have been investigated, while maintaining the desired false positive rate of $5\%$ for normal agents.
Yet, the detection method is not tailored to a specific type of anomalous behavior and only requires a data of normal robot behavior.
Additionally, while the neural network is only trained on simulations, the sensitivity and specificity results of simulated agents are consistent with the results obtained from hardware experiments.
This proven ability to train only with simulation data circumvents one of the classic main issues when applying machine learning approaches to physical engineering problems, namely the hardship to obtain enough data.

The very satisfactory performance will allow us to build upon this work while learning increasingly complex normal behavior.
For instance, it is planned to detect anomalous behavior while considering previous actions with the aim of introducing higher level goals like the evasion of obstacles.
Additionally, we will further investigate scenarios where multiple antagonists are present and continue to verify all approaches with hardware experiments. 

\section{Acknowledgements}
Many thanks to Benedict Röder, Mario Rosenfelder, Luca Jung, Jonas Kneifl, Niklas Fahse and Andreas Schönle for the helpful discussions and their aid with collecting hardware data.

The ITM acknowledges the support by the Deutsche Forschungsgemeinschaft (DFG, German Research Foundation) under Germany’s Excellence Strategy – EXC 2075 – 390740016, project PN4-4 “Learning from Data - Predictive Control in Adaptive Multi-Agent Scenarios” as well as project EB195/32-1, 433183605 “Research on Multibody Dynamics and Control for Collaborative Elastic Object Transportation by a Heterogeneous Swarm with Aerial and Land-Based Mobile Robots” and project EB195/40-1, 501890093 “Mehr Intelligenz wagen - Designassistenten in Mechanik und Dynamik (SPP 2353)".


\begin{thebibliography}{38}

\expandafter\ifx\csname natexlab\endcsname\relax\def\natexlab#1{#1}\fi
\providecommand{\url}[1]{\texttt{#1}}
\providecommand{\href}[2]{#2}
\providecommand{\path}[1]{#1}
\providecommand{\DOIprefix}{doi:}
\providecommand{\ArXivprefix}{arXiv:}
\providecommand{\URLprefix}{URL: }
\providecommand{\Pubmedprefix}{pmid:}
\providecommand{\doi}[1]{\href{http://dx.doi.org/#1}{\path{#1}}}
\providecommand{\Pubmed}[1]{\href{pmid:#1}{\path{#1}}}
\providecommand{\bibinfo}[2]{#2}
\ifx\xfnm\relax \def\xfnm[#1]{\unskip,\space#1}\fi
\bibitem[1]{ZhaoEtAl24}
\bibinfo{author}{D.~Zhao}, \bibinfo{author}{H.~Luo}, \bibinfo{author}{Y.~Tu}, \bibinfo{author}{C.~Meng}, \bibinfo{author}{T.L.~Lam},
\newblock \bibinfo{title}{Snail-inspired robotic swarms: A hybrid connector drives collective adaptation in unstructured outdoor environments},
\newblock \bibinfo{journal}{Nature Communications}, \bibinfo{volume}{15}, pp. \bibinfo{pages}{3647}, \bibinfo{year}{2024}.
\DOIprefix\doi{10.1038/s41467-024-47788-2}.
\bibitem[2]{RosenfelderEbelEberhard24}
\bibinfo{author}{M.~Rosenfelder}, \bibinfo{author}{H.~Ebel}, \bibinfo{author}{P.~Eberhard},
\newblock \bibinfo{title}{Force-based organization and control scheme for the non-prehensile cooperative transportation of objects},
\newblock \bibinfo{journal}{Robotica}, \bibinfo{volume}{42}, pp. \bibinfo{pages}{611–624}, \bibinfo{year}{2024}.
\DOIprefix\doi{10.1017/S0263574723001704}.
\bibitem[3]{SiwekEtAl23}
\bibinfo{author}{J.~Siwek}, \bibinfo{author}{P.~Żywica},
	\bibinfo{author}{P.~Siwek}, \bibinfo{author}{A.~Wójcik},
	\bibinfo{author}{W.~Woch}, \bibinfo{author}{K.~Pierzyński},
	\bibinfo{author}{K.~Dyczkowski},
\newblock \bibinfo{title}{Implementation of an {{artificially empathetic robot	swarm}}},
\newblock \bibinfo{journal}{Sensors}, \bibinfo{volume}{24}, pp. \bibinfo{pages}{242}, \bibinfo{year}{2023}.
\DOIprefix\doi{10.3390/s24010242}.
\bibitem[4]{ChenEtAl24}
\bibinfo{author}{J.~Chen}, \bibinfo{author}{W.~Luo}, \bibinfo{author}{H.~Ebel},
	\bibinfo{author}{P.~Eberhard},
\newblock \bibinfo{title}{Optimization-based trajectory planning for transport	collaboration of heterogeneous systems},
\newblock \bibinfo{journal}{at - Automatisierungstechnik}, \bibinfo{volume}{72}, pp. \bibinfo{pages}{80--90}, \bibinfo{year}{2024}.
\DOIprefix\doi{10.1515/auto-2023-0078}.
\bibitem[5]{SchranzEtAl20}
\bibinfo{author}{M.~Schranz}, \bibinfo{author}{M.~Umlauft},
	\bibinfo{author}{M.~Sende}, \bibinfo{author}{W.~Elmenreich},
\newblock \bibinfo{title}{Swarm {{robotic behaviors}} and {{current applications}}},
\newblock \bibinfo{journal}{Frontiers in Robotics and AI}, \bibinfo{volume}{7}, pp. \bibinfo{pages}{36}, \bibinfo{year}{2020}.
\DOIprefix\doi{10.3389/frobt.2020.00036}.
\bibitem[6]{SargeantTomlinson18}
\bibinfo{author}{I.~Sargeant}, \bibinfo{author}{A.~Tomlinson},
\newblock \bibinfo{title}{Review of {{potential attacks}} on {{robotic swarms}}},
\newblock in: \bibinfo{editor}{Y.~Bi}, \bibinfo{editor}{S.~Kapoor},
	\bibinfo{editor}{R.~Bhatia} (Eds.), \bibinfo{booktitle}{Proceedings of {{SAI
	Intelligent Systems Conference}} ({{IntelliSys}}) 2016}, volume~\bibinfo{volume}{16}, pp. \bibinfo{pages}{628--646}, \bibinfo{publisher}{{Springer International
	Publishing}}, \bibinfo{year}{2017}.
\DOIprefix\doi{10.1007/978-3-319-56991-8_46}.
\bibitem[7]{HigginsTomlinsonMartin09}
\bibinfo{author}{F.~Higgins}, \bibinfo{author}{A.~Tomlinson},
	\bibinfo{author}{K.M.~Martin},
\newblock \bibinfo{title}{Survey on {{security challenges}} for {{swarm
	robotics}}},
\newblock in: \bibinfo{booktitle}{Proceedings of the {{Fifth International Conference}} on
	{{Autonomic}} and {{Autonomous Systems}}},  pp.
	\bibinfo{pages}{307--312}, \bibinfo{publisher}{{IEEE}}, \bibinfo{year}{2009}. 
\DOIprefix\doi{10.1109/ICAS.2009.62}.
\bibitem[8]{SargeantTomlinson20}
\bibinfo{author}{I.~Sargeant}, \bibinfo{author}{A.~Tomlinson},
\newblock \bibinfo{title}{Intrusion {{detection}} in {{robotic swarms}}},
\newblock in: \bibinfo{editor}{Y.~Bi}, \bibinfo{editor}{R.~Bhatia},
	\bibinfo{editor}{S.~Kapoor} (Eds.), \bibinfo{booktitle}{Intelligent
	{{Systems}} and {{Applications}}}, \bibinfo{volume}{1038}, pp.
	\bibinfo{pages}{968--980}, \bibinfo{publisher}{{Springer International Publishing}}, \bibinfo{year}{2020}. 
\DOIprefix\doi{10.1007/978-3-030-29513-4_71}.
\bibitem[9]{QinHeZhou14}
\bibinfo{author}{L.~Qin}, \bibinfo{author}{X.~He}, \bibinfo{author}{D.~Zhou},
\newblock \bibinfo{title}{{A survey of fault diagnosis for swarm systems}},
\newblock \bibinfo{journal}{Systems Science \& Control Engineering}, \bibinfo{volume}{2}, pp. \bibinfo{pages}{13--23}, \bibinfo{year}{2014}.
\DOIprefix\doi{10.1080/21642583.2013.873745}.
\bibitem[10]{SaulnierEtAl17}
\bibinfo{author}{K.~Saulnier}, \bibinfo{author}{D.~Saldaña},
	\bibinfo{author}{A.~Prorok}, \bibinfo{author}{G.J.~Pappas},
	\bibinfo{author}{V.~Kumar},
\newblock \bibinfo{title}{Resilient {{flocking}} for {{mobile robot teams}}},
\newblock \bibinfo{journal}{IEEE Robotics and Automation Letters}, \bibinfo{volume}{2}, pp. \bibinfo{pages}{1039--1046}, \bibinfo{year}{2017}.
\DOIprefix\doi{10.1109/LRA.2017.2655142}.
\bibitem[11]{CavorsiEtAl22}
\bibinfo{author}{M.~Cavorsi}, \bibinfo{author}{O.E.~Akgün},
	\bibinfo{author}{M.~Yemini}, \bibinfo{author}{A.J.~Goldsmith},
	\bibinfo{author}{S.~Gil},
\newblock \bibinfo{title}{Exploiting trust for resilient hypothesis testing
	with malicious robots},
\newblock in: \bibinfo{booktitle}{2023 IEEE International Conference on
	Robotics and Automation (ICRA)},  pp. \bibinfo{pages}{7663--7669}, \bibinfo{year}{2023}. 
\DOIprefix\doi{10.1109/ICRA48891.2023.10160385}.
\bibitem[12]{DoostmohammadianElAl21}
\bibinfo{author}{M.~Doostmohammadian}, \bibinfo{author}{H.~Zarrabi},
	\bibinfo{author}{H.R.~Rabiee}, \bibinfo{author}{U.A.~Khan},
	\bibinfo{author}{T.~Charalambous},
\newblock \bibinfo{title}{Distributed detection and mitigation of biasing
	attacks over multi-agent networks},
\newblock \bibinfo{journal}{IEEE Transactions on Network Science and Engineering}, \bibinfo{volume}{8}, pp. \bibinfo{pages}{3465--3477}, \bibinfo{year}{2021}. \DOIprefix\doi{10.1109/TNSE.2021.3115032}.
\bibitem[13]{BabaeiGhazvini2023}
\bibinfo{author}{H.R.~Babaei~Ghazvini}, \bibinfo{author}{M.~Haeri},
\newblock \bibinfo{title}{Byzantine {{agents}}’ {{detection}} in {{distributed Nash equilibrium seeking algorithms using}} an {{adaptive event-triggered scheme}}},
\newblock \bibinfo{journal}{IEEE Systems Journal}, \bibinfo{volume}{17}, pp. \bibinfo{pages}{4407--4418}, \bibinfo{year}{2023}. 
\DOIprefix\doi{10.1109/JSYST.2023.3267160}.
\bibitem[14]{Zhang2021}
\bibinfo{author}{P.~Zhang}, \bibinfo{author}{C.~Hu}, \bibinfo{author}{S.~Wu},
	\bibinfo{author}{R.~Gong}, \bibinfo{author}{Z.~Luo},
\newblock \bibinfo{title}{Event-{{triggered resilient average consensus with adversary detection}} in the {{presence}} of {{Byzantine agents}}},
\newblock \bibinfo{journal}{IEEE Access}, \bibinfo{volume}{9}, pp. \bibinfo{pages}{121431--121444}, \bibinfo{year}{2021}.
\DOIprefix\doi{10.1109/ACCESS.2021.3108639}.
\bibitem[15]{Deng2021}
\bibinfo{author}{G.~Deng}, \bibinfo{author}{Y.~Zhou}, \bibinfo{author}{Y.~Xu},
	\bibinfo{author}{T.~Zhang}, \bibinfo{author}{Y.~Liu},
\newblock \bibinfo{title}{An {{investigation}} of {{Byzantine threats}} in
	{{multi-robot systems}}},
\newblock in: \bibinfo{booktitle}{24th {{International Symposium}} on
	{{Research}} in {{Attacks}}, {{Intrusions}} and {{Defenses}}}, pp.
\bibinfo{pages}{17--32}, \bibinfo{publisher}{ACM}, \bibinfo{year}{2021}. \DOIprefix\doi{10.1145/3471621.3471867}.
\bibitem[16]{BasanBasanNekrasov19}
\bibinfo{author}{E.~Basan}, \bibinfo{author}{A.~Basan},
	\bibinfo{author}{A.~Nekrasov},
\newblock \bibinfo{title}{Method for {{detecting abnormal activity}} in a {{group}} of {{mobile robots}}},
\newblock \bibinfo{journal}{Sensors},  \bibinfo{volume}{19}, pp. \bibinfo{pages}{4007}, \bibinfo{year}{2019}.
\DOIprefix\doi{10.3390/s19184007}.
\bibitem[17]{SalimpourEtAl23}
\bibinfo{author}{S.~Salimpour}, \bibinfo{author}{F.~Keramat},
	\bibinfo{author}{J.P.~Queralta}, \bibinfo{author}{T.~Westerlund},
\bibinfo{title}{Decentralized {{vision-based Byzantine agent detection}} in {{multi-robot systems}} with {{IOTA smart contracts}}},
\newblock in: \bibinfo{editor}{G.V.~Jourdan}, \bibinfo{editor}{L.~Mounier},
\bibinfo{editor}{C.~Adams}, \bibinfo{editor}{F.~S{\`e}des}, \bibinfo{editor}{J.~Garcia-Alfaro} (Eds.), \bibinfo{booktitle}{Foundations and Practice of Security}, pp. \bibinfo{pages}{322--337}, \bibinfo{publisher}{Springer Nature Switzerland}, \bibinfo{year}{2023}.
\DOIprefix\doi{10.1007/978-3-031-30122-3_20}
\bibitem[18]{WangShangSun11}
\bibinfo{author}{C.~Wang}, \bibinfo{author}{W.~Shang},
	\bibinfo{author}{D.~Sun},
\newblock \bibinfo{title}{Monitoring malfunction in multirobot formation with a
neural network detector},
\newblock \bibinfo{journal}{Proceedings of the Institution of Mechanical Engineers, Part I: Journal of Systems and Control Engineering}, \bibinfo{volume}{225}, pp. \bibinfo{pages}{1163--1172}, \bibinfo{year}{2011}. 
\DOIprefix\doi{10.1177/0959651811400530}.
\bibitem[19]{ChandolaBanerjeeKumar09}
\bibinfo{author}{V.~Chandola}, \bibinfo{author}{A.~Banerjee},
	\bibinfo{author}{V.~Kumar},
\newblock \bibinfo{title}{Anomaly detection: a survey},
\newblock \bibinfo{journal}{ACM Computing Surveys},  \bibinfo{volume}{41}, pp.
\bibinfo{pages}{1--58}, \bibinfo{year}{2009}.
\DOIprefix\doi{10.1145/1541880.1541882}.
\bibitem[20]{FagioliniEtAl09}
\bibinfo{author}{A.~Fagiolini}, \bibinfo{author}{F.~Babboni},
	\bibinfo{author}{A.~Bicchi},
\newblock \bibinfo{title}{Dynamic {{distributed intrusion detection}} for {{secure multi-robot systems}}},
\newblock in: \bibinfo{booktitle}{2009 {{IEEE International Conference}} on {{Robotics}} and {{Automation}}},  pp. \bibinfo{pages}{2723--2728}, \bibinfo{publisher}{IEEE}, \bibinfo{year}{2009}.
\DOIprefix\doi{10.1109/ROBOT.2009.5152608}.
\bibitem[21]{WengerEbelEberhard24}
\bibinfo{author}{I.~Wenger}, \bibinfo{author}{H.~Ebel},
	\bibinfo{author}{P.~Eberhard},
\newblock \bibinfo{title}{{{Anomalously acting agents: the deployment problem}}},
\newblock \bibinfo{journal}{Multibody System Dynamics},	pp. \bibinfo{pages}{1--19}, \bibinfo{year}{2024}. 
\DOIprefix\doi{10.1007/s11044-024-09993-1}.
\bibitem[22]{CandidoHutchinson09}
\bibinfo{author}{S.~Candido}, \bibinfo{author}{S.~Hutchinson},
\newblock \bibinfo{title}{Detecting {intrusion faults} in {remotely controlled
	systems}},
\newblock in: \bibinfo{booktitle}{{{Proceedings of the 2009 American Control Conference}}}, pp. \bibinfo{pages}{4968--4973}, \bibinfo{publisher}{IEEE}, \bibinfo{year}{2009}.
\DOIprefix\doi{10.1109/ACC.2009.5160086}.
\bibitem[23]{ZhouEtAl24}
\bibinfo{author}{M.~Zhou}, \bibinfo{author}{J.~Li}, \bibinfo{author}{C.~Wang},
	\bibinfo{author}{J.~Wang}, \bibinfo{author}{L.~Wang},
\newblock \bibinfo{title}{Applications of {{Voronoi diagrams}} in {{multi-robot
	coverage}}: {{a review}}},
\newblock \bibinfo{journal}{Journal of Marine Science and Engineering}, \bibinfo{volume}{12}, pp. \bibinfo{pages}{1022}, \bibinfo{year}{2024}. 
\DOIprefix\doi{10.3390/jmse12061022}.
\bibitem[24]{Lloyd82}
\bibinfo{author}{S.~Lloyd},
\newblock \bibinfo{title}{Least squares quantization in {PCM}},
\newblock \bibinfo{journal}{IEEE Transactions on Information Theory}, \bibinfo{volume}{28}, pp. \bibinfo{pages}{129--137}, \bibinfo{year}{1982}.
\DOIprefix\doi{10.1109/TIT.1982.1056489}.
\bibitem[25]{Voronoi08}
\bibinfo{author}{G.~Voronoi},
\newblock \bibinfo{title}{Nouvelles applications des param{\`e}tres continus
	{\`a} la th{\'e}orie des formes quadratiques. Premier m{\'e}moire. Sur
	quelques propri{\'e}t{\'e}s des formes quadratiques positives parfaites},
\newblock \bibinfo{journal}{Journal f{\"u}r die reine und angewandte Mathematik
	(Crelles Journal)}, \bibinfo{volume}{1908},	pp. \bibinfo{pages}{97--102}, \bibinfo{year}{1908}. 
\DOIprefix\doi{10.1515/crll.1908.133.97}.
\bibitem[26]{JeffreyTanVillar23}
\bibinfo{author}{N.~Jeffrey}, \bibinfo{author}{Q.~Tan}, \bibinfo{author}{J.~R.
	Villar},
\newblock \bibinfo{title}{A {{review}} of {{anomaly detection strategies}} to
	{{detect threats}} to {{cyber-physical systems}}}, 
\newblock \bibinfo{journal}{Electronics}, \bibinfo{volume}{12}, pp. \bibinfo{pages}{3283}, \bibinfo{year}{2023}.
\DOIprefix\doi{10.3390/electronics12153283}.
\bibitem[27]{HochreiterSchmidhuber97}
\bibinfo{author}{S.~Hochreiter}, \bibinfo{author}{J.~Schmidhuber},
\newblock \bibinfo{title}{Long short-term memory},
\newblock \bibinfo{journal}{Neural Computation}, \bibinfo{volume}{9}, pp. \bibinfo{pages}{1735–1780}, \bibinfo{year}{1997}.
\DOIprefix\doi{10.1162/neco.1997.9.8.1735}.
\bibitem[28]{SchusterPaliwal97}
\bibinfo{author}{M.~Schuster}, \bibinfo{author}{K.~Paliwal},
\newblock \bibinfo{title}{Bidirectional recurrent neural networks},
\newblock \bibinfo{journal}{IEEE Transactions on Signal Processing}, \bibinfo{volume}{45}, pp. \bibinfo{pages}{2673--2681}, \bibinfo{year}{1997}.
\DOIprefix\doi{10.1109/78.650093}.
\bibitem[29]{Polya20}
\bibinfo{author}{G.~P{\'o}lya},
\newblock \bibinfo{title}{{\"U}ber den zentralen {G}renzwertsatz der
	{W}ahrscheinlichkeits-rechnung und das {M}omentenproblem},
\newblock \bibinfo{journal}{Mathematische Zeitschrift}, \bibinfo{volume}{8}, pp. \bibinfo{pages}{171--181}, \bibinfo{year}{1920}.
\DOIprefix\doi{10.1007/BF01206525}.
\bibitem[30]{TabakTurner13}
\bibinfo{author}{E.G.~Tabak}, \bibinfo{author}{C.V.~Turner},
\newblock \bibinfo{title}{A {{family}} of {{nonparametric density estimation
	algorithms}}},
\newblock \bibinfo{journal}{Communications on Pure and Applied Mathematics}, \bibinfo{volume}{66}, pp. \bibinfo{pages}{145--164}, \bibinfo{year}{2013}.
	\DOIprefix\doi{10.1002/cpa.21423}.
\bibitem[31]{RezendeMohamed15}
\bibinfo{author}{D.J.~Rezende}, \bibinfo{author}{S.~Mohamed},
\newblock \bibinfo{title}{Variational {inference} with {normalizing flows}},
\newblock in: \bibinfo{editor}{F.~Bach}, \bibinfo{editor}{D.~Blei} (Eds.),
\bibinfo{booktitle}{Proceedings of the 32nd International Conference on Machine Learning}, \bibinfo{volume}{37}, pp. \bibinfo{pages}{1530--1538}, \bibinfo{publisher}{PMLR}, \bibinfo{year}{2015}. 
\DOIprefix\doi{10.5555/3045118.3045281}.
\bibitem[32]{PapamakariosEtAl21}
\bibinfo{author}{G.~Papamakarios}, \bibinfo{author}{E.~Nalisnick},
	\bibinfo{author}{D.J.~Rezende}, \bibinfo{author}{S.~Mohamed},
	\bibinfo{author}{B.~Lakshminarayanan},
\newblock \bibinfo{title}{Normalizing {{flows}} for {{probabilistic modeling}}
	and {{inference}}},
\newblock \bibinfo{journal}{The Journal of Machine Learning Research},	\bibinfo{volume}{22}, pp. \bibinfo{pages}{2617--2680}, \bibinfo{year}{2021}.
\DOIprefix\doi{10.48550/arXiv.1912.02762}.
\bibitem[33]{nflows}
\bibinfo{author}{C.~Durkan}, \bibinfo{author}{A.~Bekasov},
	\bibinfo{author}{I.~Murray}, \bibinfo{author}{G.~Papamakarios},
	\bibinfo{title}{{nflows}: normalizing flows in {PyTorch}},
	\bibinfo{howpublished}{Zenodo},  \bibinfo{year}{2020}.
\DOIprefix\doi{10.5281/zenodo.4296287}.
\bibitem[34]{HuangOlsonMoore10}
\bibinfo{author}{A.S.~Huang}, \bibinfo{author}{E.~Olson},
	\bibinfo{author}{D.C.~Moore},
\newblock \bibinfo{title}{{LCM}: Lightweight communications and marshalling},
\newblock in: \bibinfo{booktitle}{Proceedings of the 2010 IEEE/RSJ
	International Conference on Intelligent Robots and Systems}, pp. \bibinfo{pages}{4057--4062}, \bibinfo{publisher}{IEEE}, \bibinfo{year}{2010}. 
\DOIprefix\doi{10.1109/IROS.2010.5649358}.
\bibitem[35]{Ebel21}
\bibinfo{author}{H.~Ebel},
\newblock \bibinfo{title}{Distributed control and organization of communicating mobile robots: Design, simulation, and experimentation},
\bibinfo{series}{Dissertation, Schriften aus dem Institut f\"{u}r Technische und Numerische Mechanik der Universit\"{a}t Stuttgart}, \bibinfo{volume}{69}.
\newblock \bibinfo{publisher}{Shaker Verlag}, \bibinfo{address}{D{\"u}ren}, \bibinfo{year}{2021}.
\DOIprefix\doi{http://dx.doi.org/10.18419/opus-13315}.
\bibitem[36]{PaszkeEtAl19}
\bibinfo{author}{A.~Paszke}, \bibinfo{author}{S.~Gross},
	\bibinfo{author}{F.~Massa}, \bibinfo{author}{A.~Lerer},
	\bibinfo{author}{J.~Bradbury}, \bibinfo{author}{G.~Chanan},
	\bibinfo{author}{T.~Killeen}, \bibinfo{author}{Z.~Lin},
	\bibinfo{author}{N.~Gimelshein}, \bibinfo{author}{L.~Antiga},
	\bibinfo{author}{A.~Desmaison}, \bibinfo{author}{A.~Kopf},
	\bibinfo{author}{E.~Yang}, \bibinfo{author}{Z.~DeVito},
	\bibinfo{author}{M.~Raison}, \bibinfo{author}{A.~Tejani},
	\bibinfo{author}{S.~Chilamkurthy}, \bibinfo{author}{B.~Steiner},
	\bibinfo{author}{L.~Fang}, \bibinfo{author}{J.~Bai},
	\bibinfo{author}{S.~Chintala},
\newblock \bibinfo{title}{Pytorch: An imperative style, high-performance deep
	learning library},
\newblock in: \bibinfo{editor}{H.~Wallach}, \bibinfo{editor}{H.~Larochelle},
	\bibinfo{editor}{A.~Beygelzimer}, \bibinfo{editor}{F.~d\textquotesingle
	Alch\'{e}-Buc}, \bibinfo{editor}{E.~Fox}, \bibinfo{editor}{R.~Garnett}
	(Eds.), \bibinfo{booktitle}{Advances in Neural Information Processing Systems
	33}, pp. \bibinfo{pages}{8024--8035}, \bibinfo{publisher}{Curran Associates}, \bibinfo{year}{2019}.
	\DOIprefix\doi{10.5555/3454287.3455008}.
\bibitem[37]{EbelEberhard21}
\bibinfo{author}{E.~Ebel}, \bibinfo{author}{E.~Eberhard},
\newblock \bibinfo{title}{A comparative look at two formation control approaches based on optimization and algebraic graph theory},
\newblock \bibinfo{journal}{Robotics and Autonomous Systems}, \bibinfo{volume}{136}, pp. \bibinfo{pages}{103686}, \bibinfo{year}{2021}.
\DOIprefix\doi{10.1016/j.robot.2020.103686}.
\bibitem[38]{wandb}
\bibinfo{author}{L.~Biewald}, \bibinfo{title}{Experiment tracking with weights
	and biases}, \bibinfo{url}{https://pypi.org/project/wandb/}, \bibinfo{year}{2020}, 0.16.1.
\bibitem[39]{BrunnerMunzel00}
\bibinfo{author}{E.~Brunner}, \bibinfo{author}{U.~Munzel},
\newblock \bibinfo{title}{The {{nonparametric Behrens-Fisher problem}}:
	{{Asymptotic theory}} and a {{small-sample approximation}}},
\newblock \bibinfo{journal}{Biometrical Journal}, \bibinfo{volume}{42}, pp. \bibinfo{pages}{17--25}, \bibinfo{year}{2000}.
\DOIprefix\doi{10.1002/(SICI)1521-4036(200001)42:1$<$17::AID-BIMJ17$>$3.0.CO;2-U}.
	
\end{thebibliography}
\end{document}